\documentclass[review,11.5pt]{elsarticle}
\usepackage{lineno}
\usepackage{booktabs}
\usepackage{float}
\usepackage{amsmath}
\usepackage{amssymb}
\usepackage{adjustbox}
\usepackage{graphicx}
\usepackage[table,xcdraw]{xcolor}
\usepackage{verbatim}
\usepackage{pdflscape}
\usepackage{hhline}
\usepackage{subfigure}
\usepackage{threeparttable}
\journal{Decision Support Systems}

\begin{document}

\begin{frontmatter}

\title{An empirical comparison of deep-neural-network architectures for next activity prediction using context-enriched process event logs} 


\author[firstaddress]{Sven Weinzierl\corref{mycorrespondingauthor}}
\cortext[mycorrespondingauthor]{Corresponding authors}
\ead{sven.weinzierl@fau.de}
\author[firstaddress]{Sandra Zilker}
\author[secondaddress]{Jens Brunk}
\author[thirdaddress]{Kate Revoredo}
\author[fourthaddress]{An Nguyen}
\author[firstaddress]{Martin Matzner}
\author[secondaddress]{J{\"o}rg Becker}
\author[fourthaddress]{Bj{\"o}rn Eskofier}
\address[firstaddress]{
Institute of Information Systems, Friedrich-Alexander-Universit\"at Erlangen-N{\"u}rnberg, F{\"u}rther Stra{\ss}e 248, Nuremberg, Germany}

\address[secondaddress]{University of M{\"u}nster - ERCIS, Leonardo-Campus 3, M{\"u}nster, Germany}

\address[thirdaddress]{Department of Information Systems and Operations, Vienna University of Economics and Business (WU), Vienna, Austria}

\address[fourthaddress]{Department of Computer Science, Friedrich-Alexander-Universit\"at 
Erlangen-N{\"u}rnberg, Carl-Thiersch-Stra{\ss}e 2b, Erlangen, Germany}

\begin{abstract}
%
Researchers have proposed a variety of predictive business process monitoring (PBPM) techniques aiming to predict future process behaviour during the process execution. 
Especially, techniques for the next activity prediction anticipate great potential in improving operational business processes.
To gain more accurate predictions, a plethora of these techniques rely on deep neural networks (DNNs) and consider information about the context, in which the process is running.
However, an in-depth comparison of such techniques is missing in the PBPM literature, which prevents researchers and practitioners from selecting the best solution for a given event log.
To remedy this problem, we empirically evaluate the predictive quality of three promising DNN architectures, combined with five proven encoding techniques 
and based on five context-enriched real-life event logs.
%
We provide four findings that can support researchers and practitioners in designing novel PBPM techniques for predicting the next activities.
\end{abstract}
\begin{keyword}
Business process management, predictive business process monitoring, deep learning, process context.
\end{keyword}
\end{frontmatter}


\section{Introduction}
The highly volatile and uncertain digital economy increases the pressure on organisations to immediately improve their business processes \citep{poll.2018}.
As a consequence, predictive business process monitoring (PBPM) is gaining momentum in business process management (BPM)~\citep{breuker.2016}. 
PBPM provides a set of prediction techniques to improve operational business processes.
A PBPM technique aims to predict the future process behaviour during the process execution~\citep{maggi.2014,schwegmann-2013-method-tool} by using predictive models constructed from historical process event data~\citep{marquez.2017a}.
Most of the recent PBPM techniques use machine-learning (ML) algorithms to learn predictive models. Generally, a predictive model is geared to a prediction task~\citep{di.2018, marquez.2017a}. This can be, for instance, next activity predictions~\citep{tama.2019}, process outcome predictions~\citep{teinemaa.2019} or remaining time predictions~\citep{verenich.2019}.  

In PBPM, the next activity prediction is one of the most researched tasks~\citep{di.2018}. 
It aims to anticipate future process steps in the course of a running process instance. 
The task's importance is emphasised, among others, by four analytical uses:
(1)~\textit{Early warning} -- Monitoring the next most likely activities to be proactively aware of inefficiencies, risks, mistakes~\citep{di.2017} or complexities~\citep{weinzierl.2020}. 
%
(2)~\textit{Best action recommendation} -- Determining the next best actions based on correctly predicted (next most likely) activities and depending on desired key performance indicators~\citep{weinzierl.2020b}. 
(3)~\textit{Anomaly detection} -- Identifying anomalous process instances by considering the probability distribution of next activity predictions~\citep{nolle.2019}. 
%
(4)~\textit{Resource allocation} -- Allocating resources proactively based on next activity predictions~\citep{park.2019}. 
%

Deep learning (DL), a subarea of ML, has found its way into PBPM through the next activity prediction in the form of deep neural networks (DNN)~\citep{evermann.2017}.
%
%
%
Generally, DNNs learn a more abstract, hierarchical representation of the data through multiple hidden layers to grasp the intricate structure in data~\citep{lecun.2015}. 
Therefore, concerning the next activity prediction, DNNs can learn predictive models more accurately \citep{evermann.2017} since they better take the more complex sequential structure of event log data into account.
So far, three different types of DL approaches have been successfully applied for the next activity prediction. 
%
These architectural types are: 
(1) multi-layer perceptron (MLP)~\citep{mehdiyev.2018,theis.2019}, 
(2) recurrent neural network (RNN)~(e.g. \citep{evermann.2017,tax.2017,weinzierl.2020})
and
(3) convolutional neural network (CNN)~\citep{al.2018, pasquadibisceglie.2019, dimauro.2019}.
%

Additionally, an event log can include information about a business process's context in the form of context attributes. 
It can be defined as the "minimum set of variables containing all relevant information that impact the design[, implementation] and execution of a business process"~\citep[p.154]{rosemann.2006}.
Context information, originating from sources external~\citep{yeshchenko.2018, weinzierl.2019} or internal~(e.g. \citep{schoenig.2018, evermann.2017}) to the business process, can improve process predictions since it adds valuable information to the predictive models~\citep{marquez.2017a}.
In next activity predictions, context information might affect the future direction in the course of a running process instance~(e.g. \citep{schoenig.2018, evermann.2017}). 
A standard DNN requires an appropriate encoding of the context attributes to benefit from context information~\citep{weinzierl.2020, evermann.2017, camargo.2019}.
Otherwise, valuable information about the context remains hidden in the data attributes, and a DNN learns models with a limited predictive quality since the loss of the potential context information is high.
For the next activity prediction, Weinzierl et al.~\citep{weinzierl.2020} confirm a significant relationship between 
the representation of an event log's context attributes and a DNN's predictive quality. 
Further, we assume an impact of the representation of an event log's control-flow attributes, especially the activity attributes, on the DNN's predictive quality.   
%
To cope with information loss, researchers in the field of ML have suggested different encoding techniques for DNNs~\citep{potdar.2017}.
%
Therefore, for practitioners as well as researchers, not only the selection of the right type of DNN architecture is a challenge, but also the choice of a suitable representation of an event log's attributes. 

However, a comparison of the four proven DNN architecture types in combination with established ML encoding techniques for representing an event log's attributes is missing in the PBPM literature for the next activity prediction in the presence of context information.
%
In this paper, we overcome this challenge. 
For that, we empirically evaluate the predictive quality for an instance of each of the three applied DNN architecture types 
(MLP by Theis and Darabi~\citep{theis.2019}, 
long short-term neural network (LSTM) by Camargo et al.~\citep{camargo.2019} and 
CNN by Al-Jebrni  et  al.~\citep{al.2018}).
Additionally, we include five proven encoding techniques from the ML literature 
(binary, 
ordinal,
onehot, 
hash and 
word2vec) in our evaluation.
We base our evaluation on five context-enriched real-life event logs with different characteristics. 
With our paper, we contribute to academia and practice in three ways:
First, we give an overview of existing DL-based PBPM techniques for the next activity prediction. 
Second, we present findings that support the design of novel PBPM techniques building on DNNs using context for the next activity prediction.
Third, we show directions for future research concerning the next activity prediction with DNNs using context. 

The remainder of this paper is structured as follows: 
Section \ref{sec:background} presents the preliminaries and related work on the DL-based next activity prediction. 
Section \ref{sec:setting} describes 
the experimental setting of our work.
%
In Section \ref{sec:results}, we present and subsequently discuss the experimental results in Section \ref{sec:discussion}. 
Section \ref{sec:conclusion} concludes our work and points to future research directions.

\section{Background}
\label{sec:background}
\subsection{Preliminaries}
\label{sec:preliminaries}
PBPM techniques construct predictive models from 
%
event log data. An \emph{event log} includes detailed information about the activities that have been executed in a single process~\citep{van.2016}.
%
We adapt definitions by Polato et al.~\citep{polato.2014} to formally describe the terms: \textit{event universe}, \textit{event}, \textit{trace}, \textit{event log} and \textit{next activity}. 
First, an \emph{event universe} defines all possible events in an event log.

%
\noindent
\textbf{Definition 1 (Event universe).}
$\mathcal{A}$ is the set of process activities, $\mathcal{C}$ is the set of process instances~(cases), $C$ is the set of case ids with the bijective projection $id : C \to \mathcal{C}$, $\mathcal{T}$ is the set of timestamps and $\mathcal{D}$ is the set of possible context attributes. 
To address time, a process instance $c \in \mathcal{C}$ contains all past and future events, while events in a trace $\sigma_c$ of $c$ contain all events up to the currently available time instant.
$\mathcal{E}=\mathcal{A} \times C \times \mathcal{T} \times \mathcal{D}$ is the event universe.

An \textit{event} is the specific record of the execution of an activity. It contains information on activity, case id, timestamp and context attributes.
\\
\noindent
\textbf{Definition 2 (Event).}
  An event $e \in \mathcal{E}$
  is a tuple $e=(a,c,t,D)$,
  where $a \in \mathcal{A}$
  is the process activity,
  $c \in C$ is the case id, $t \in \mathcal{T}$ is its start timestamp and $D \subseteq \mathcal{D}$ is a set of context attributes. $d \in D$ is a context attribute of event $e$.
  %
  Given an event $e$,
  we define the projection functions
  $F_{p_{e}}=\{f_{a}, f_{c}, f_{t}, f_{D}\}$: $f_{a}: e \to a, f_{c}: e \to c, f_{t}: e \to t$ and $f_{D}: e \to D$.
  Given a context attribute $d$, we define the projection functions\footnote{Note ``ec" = ``event-related context attribute", ``pc" = ``process-related context attribute", ``num" = ``numerical context attribute" and ``cat" = ``categorical context attribute".} $F_{p_{d}}=\{f_{ec}, f_{pc}, f_{num}, f_{cat}\}: f_{ec}: d \to \mathcal{D}_{ec}, f_{pc}: d \to \mathcal{D}_{pc}, f_{num}: d \to \mathcal{D}_{num}$ and $f_{cat}: d \to \mathcal{D}_{cat}.$ 

A \emph{trace} includes the stored information on an instance of a business process (i.e. a specific execution of a business process) in an event log. 
\\
\noindent
\textbf{Definition 3 (Trace).}
  A trace is a 
  sequence $\sigma_{c} = \langle e_{1}, \dots, e_{\vert \sigma_{c} \vert} \rangle \in \mathcal{E}^*$ of events, such that $f_{c}(e_{i}) = f_{c}(e_{j}) \wedge f_{t}(e_i) \leq f_{t}(e_j)$ for $1 \leq i < j \leq \vert \sigma_{c} \vert$. 
  Given $\sigma_{c}= \langle e_{1},..,e_{k},.., \vert \sigma_{c} \vert \rangle$,
  the prefix of length $k$ 
  is defined as $hd^k(\sigma_{c})= \langle e_{1},..,e_{k} \rangle$, with $0 < k < \vert \sigma_{c} \vert$.
\\
An \emph{event log} is a collection of past instances of a business process.
\\
\noindent
\textbf{Definition 4 (Event log).}
  An event log $\mathcal{L}_\tau$
  for a time instant $\tau$
  is a set of traces,
  such that for all $\sigma{}_c \in \mathcal{L}_\tau\,
  $ it exists a case
  $c \in \mathcal{C}\,$ with
  $\, ($for all $e \in \sigma_c\,$ it holds
  $\,id(f_c(e)) = c)$ and $($for all $e \in \sigma_c\,$ it holds $\,f_t(e) \leq \tau)$,
  i.e. all events of the observed cases that already happened.
\\
Based on event log data, we refer in this paper to the next activity prediction.
\\
\noindent
\textbf{Definition 5 (Next activity).}
Given a time instant $t$ ($t>\tau$), at that a running trace $\sigma_{c}^r$ has a length of $k$ $\langle e_{1},\dots,e_{k}\rangle$, the next activity of $\sigma_{c}^r$ is $f_{a}(next(\sigma_{c}^r)=e_{k+1}))$, with $0 < k < \vert \sigma_{c}^r \vert$. $e_{\vert \sigma_{c}^r \vert}\in \sigma_{c}^r$ ends $\sigma_{c}^r$. 
%

\subsection{Deep-Learning-based PBPM techniques for next activity prediction}
\label{sec:dl_techniques}
Table~\ref{tab:relatedwork} summarises works proposing different DNN architectures and encoding techniques for the next activity prediction.
It groups the works by the DNN architecture types: 
(1) \textit{MLP}, 
(2) \textit{RNN}, 
(3) \textit{CNN} and  
(4) \textit{Others}. 
The type \textit{Others} includes DNN architectures, which are 
generally not suitable for the next activity prediction according to our ``next activity" definition in Section \ref{sec:preliminaries}. 
Additionally, for each work, it reports whether context attributes are considered and the encoding techniques applied for the control-flow attributes (i.e. activity and time) as well as the context attributes (i.e. numerical and categorical attributes).
In the following, we present the works of each type and choose one DNN architecture from type (1) to (3) for our evaluation. 
\begin{table}[!htbp]
\caption{Related work on deep-learning architectures and encoding techniques.}
\label{tab:relatedwork}
\begin{adjustbox}{max width=\textwidth}
\centering
\begin{tabular}{|l|l|l|c|l|p{2.2cm}|p{2.2cm}|p{2.2cm}|p{2.2cm}|} 
\cline{5-8}
\multicolumn{4}{l|}{}  &                                \multicolumn{4}{c|}{\textbf{Encoding techniques}}  \\ 
\cline{5-8}
\multicolumn{4}{l|}{} & \multicolumn{2}{c|}{\textbf{Control-flow attributes}} & \multicolumn{2}{c|}{\textbf{Context attributes}}  \\ 
\hline
\textbf{Arch. type} & 
\textbf{Paper} & 
\textbf{Architecture}& 
\textbf{Context}&
\textbf{Activity} & 
\textbf{Time} & 
\textbf{Numerical}    & 
\textbf{Categorical}                
\\ 
\hline
(1) MLP
& 
Theis and Darabi~\citep{theis.2019}& 
MLP             & 
x&
Standardisation  & 
-                           & 
Discretisation; stand.      & 
Stand. 
\\
\hhline{~|*7{-}|}
& Mehdiyev et al.~\citep{mehdiyev.2018}            
& MLP + AE       &       
& Hash     
& --                                
& Not reported & Not reported
\\
\hline 
(2) RNN
&
\begin{tabular}[l]{@{}l@{}}Tax et al.~\citep{tax.2017}; \ Park and Song~\citep{park.2019};\\Di Francescomarino et al.~\citep{di.2017}\end{tabular}
&
LSTM & 
&
\begin{tabular}[l]{@{}l@{}}Onehot; ordinal\end{tabular} & 
** & 
-- & 
--
\\
\hhline{~|*7{-}|}
& 
Evermann et al.~\citep{evermann.2017} &
LSTM            & 
x&
Embedding &
-- & 
-- & 
Embedding 
\\
\hhline{~|*7{-}|}
& 
Sch\"onig et al.~\citep{schoenig.2018} &
LSTM            & 
x&
Onehot  &
-- & 
Min-max      &  
Onehot 
\\
\hhline{~|*7{-}|}
& 
Tello-Leal et al.~\citep{tello.2018}  & 
LSTM            &
x&
Embedding  & 
-- &
-- & 
Embedding 
\\
\hhline{~|*7{-}|}
& 
Camargo et al.~\citep{camargo.2019} & 
LSTM &
x&
Embedding &
Min-max; log  &
-- & 
Embedding 
\\
\hhline{~|*7{-}|}
& 
Weinzierl et al.~\citep{weinzierl.2020} &
LSTM  & 
x&
Onehot & 
-- & 
-- & 
Doc2vec  
\\
\hhline{~|*7{-}|}
&
Weinzierl et al.~\citep{weinzierl.2020b}&
LSTM &
x&
\begin{tabular}[l]{@{}l@{}}Onehot; ordinal\end{tabular} & 
** & 
-- & 
Ordinal
\\
\hhline{~|*7{-}|}
&
Metzger et al.~\citep{metzger.2019}
&
LSTM  & 
x&
Onehot & 
Not reported & 
Not reported & 
Not reported  
\\
\hhline{~|*7{-}|}
& 
Hinkka et al.~\citep{hinkka.2019} & 
GRU & 
x&
Onehot$^*$ & 
-- & 
-- & 
Onehot$^*$
\\
\hhline{~|*7{-}|}
& 
Nolle et al.~\citep{nolle.2019}  & 
GRU &
x&
Embedding & 
-- & 
Embedding & 
Embedding
\\
\hline
(3) CNN 
&
Al-Jebrni et al.~\citep{al.2018}  &
CNN             & &
Embedding &
--                                &
--           &
--
\\
\hhline{~|*7{-}|}
&
Pasquadibisceglie et al.~\citep{pasquadibisceglie.2019}   & 
CNN             & &
Onehot accu.  & 
Not encoded & 
--           & 
-- 
\\
\hhline{~|*7{-}|}
& 
Di Mauro et al.~\citep{dimauro.2019} &
CNN             & &
Embedding &
Not encoded &
--           &
--
\\ 
\hline
(4) Others           
&  
Lin et al.~\citep{lin.2019} & 
Encoder-Decoder & 
x&
Embedding & 
-- & 
-- & 
Embedding
\\
\hhline{~|*7{-}|}
& Khan et al.~\citep{khan.2018}     
& Encoder-Decoder  &           
& Embedding 
& --                                
& --           
& --                        
\\
\hline
\multicolumn{8}{l}{}$^*$ New encoding technique based on onehot encoding and clustering; $^{**}$ Four time-based attributes proposed by~Tax et al.~\citep{tax.2017} with different encodings.
\end{tabular}
\end{adjustbox}
\end{table}
\vspace{-0.5cm}
%

Theis and Darabi ~\citep{theis.2019} present
\textit{DREAM-NAPr}. 
%
%
\textit{DREAM} extracts attribute vectors by replaying an event log on a Petri net model with decay functions.  
%
%
Attributes are decay function values, token counts, Petri net markings and context attribute counts at time $t$.  
%
Continuous attributes are discretised, and then all attributes are normalised via standardisation.
\textit{NAPr} is a MLP that predicts next activities.
%
Further, Mehdiyev et al.~\citep{mehdiyev.2018}'s technique consists of two stacked autoencoders (AEs) followed by an MLP.
%
%
%
The stacked AEs pre-trains the process's representation in the latent space.
The MLP consists of two hidden layers and predicts the next activities.
We include the MLP of Theis and Darabi~\citep{theis.2019} in our evaluation since the MLP of Mehdiyev et al.~\citep{mehdiyev.2018} depends on the pre-trained weights of the two stacked AEs.
%

Moreover, most of the works~\citep{tax.2017, di.2017, park.2019, schoenig.2018, evermann.2017, tello.2018, camargo.2019, weinzierl.2020, weinzierl.2020b, metzger.2019} propose RNNs with LSTM cells for the next activity prediction~\citep{hochreiter.1997}. 
%
Three of these works \citep{tax.2017,di.2017, park.2019} do not consider context information. 
 Tax et al.~\citep{tax.2017} present a multi-task DNN architecture with three LSTM layers. The first LSTM layer works as a ``shared layer" and is connected with the second and the third LSTM layer, respectively. While the second LSTM layer refers to the next activity (event type) prediction, the third concerns the next timestamp (i.e. time to the next event) prediction.
The authors onehot encode activities. The resulting attribute vector is extended by five additional control-flow attributes. One attribute is the event's index (ordinal encoded). The others are temporal attributes with different encodings. 
%
%
The authors also predict the suffix (i.e. the remaining sequence) of the next activities and its remaining cycle time by repeatedly predicting next activities and their next timestamps until process instance completion.
Further, based on the suffix prediction of Tax et al.~\citep{tax.2017}, Di Francescomarino et al.~\citep{di.2017} propose rule-based techniques to improve the predictive quality by considering additional domain knowledge. 
Park and Song~\citep{park.2019} present a technique for reallocating resources proactively also based on the DNN architecture of Tax et al.~\citep{tax.2017}. 
%
On the contrary, seven works~\citep{evermann.2017, schoenig.2018, tello.2018, camargo.2019, weinzierl.2020, weinzierl.2020b, metzger.2019} rely on RNNs with LSTM cells considering context information.
Evermann et al.~\citep{evermann.2017} present a DNN architecture with an embedding layer\footnote{Note embedding is not the same as word2vec. Embedding is part of a DNN architecture. Thus, its vectors' values are adjusted regarding the next activities (supervised learning). It aims to reduce the dimensionality of onehot encoded vectors. In contrast, word2vec is not a DNN architecture. So, its vectors' values are not adjusted regarding the next activities (unsupervised learning). It aims to find (semantic and syntactic) similar word clouds.} followed by two LSTM layers to predict the next activities. 
For each event log, they feed into their DNN architecture one selected, categorical context attribute.  
%
Sch\"onig et al.~\citep{schoenig.2018} also propose a DNN architecture with two LSTM layers for predicting the next activity in combination with its resource. They evaluate their DNN architecture based on one event log and include one categorical and two numerical context attributes. They onehot encode all categorical attributes, whereas numerical context attributes are min-max normalised.  
Tello-Leal et al.~\citep{tello.2018}~present a DNN architecture with one LSTM layer. 
They use one event log for evaluation and include the resource attribute as context. Activities and context are onehot encoded

Camargo et al.~\citep{camargo.2019}~propose a multi-task DNN architecture with four LSTM layers for predicting next activities, their timestamps and related resource attribute values. 
First, activities and resources are fed into separate embedding layers. Then, the architecture concatenates the output of both embedding layers and the times (accumulated in the course of a process instance) and transfers the result to the first LSTM layer. The time values are min-max or log normalised. The first LSTM is followed by the other three LSTM layers. Each LSTM layer refers to one of the three prediction tasks.    
Weinzierl et al.~\citep{weinzierl.2020}~present a DNN architecture with one LSTM layer. The authors evaluate their architecture based on a log from the web mining community. They onehot encode activities and apply doc2vec~\citep{le.2014} to categorical context attributes.
%
Weinzierl et al.~\citep{weinzierl.2020b}~propose a technique to transform the next activity predictions into the next best actions depending on a KPI. The authors use a context-sensitive variant of Tax et al.~\citep{tax.2017}'s LSTM architecture. They include one categorical context attribute with an ordinal encoding.       
Metzger et al.~\citep{metzger.2019}~extend the multi-task DNN architecture of Tax et al.~\citep{tax.2017} by a (binary) process outcome prediction. They onehot encode activities. Further attributes are not reported. 
%
Hinkka et al.~\citep{hinkka.2019} rely on RNNs with gated recurrent units (GRU)~\citep{cho.2014}. 
In contrast to an LSTM cell, a GRU cell has a less complex structure. 
%
Regarding the activity and categorical context attributes, the authors report results for four attribute settings: \textit{None} (process instance without event attributes), \textit{Raw} (each attribute onehot encoded), \textit{ClustN} (onehot encoded cluster labels of event attributes clustered into $N$ clusters) and \textit{Both} (\textit{Raw} concatenated with \textit{ClustN}).   
Nolle et al.~\citep{nolle.2019} present \textit{BINet}, an approach for detecting anomalous process instances, that is based on the probability distribution of next activity (and its assigned context attributes) predictions. 
They propose a multi-task architecture with GRU layers. Its structure depends on the number of context attributes available in a given event log. 
%
%
From the RNN architecture type, we include the LSTM architecture of~Camargo et al.~\citep{camargo.2019} with two adaptations in our evaluation.
We prefer a standard LSTM architecture since variants of it like GRU (cf. Hinkka et al.~\citep{hinkka.2019} and Nolle et al.~\citep{nolle.2019}) 
do not significantly improve the predictive quality but are geared to other characteristics like a shorter training time~\citep{greff.2016}.
Tello-Leal et al.~\citep{tello.2018} and Weinzierl et al.~\citep{weinzierl.2020} predict next activities in other contexts (i.e. internet of things and web usage mining). 
Sch{\"o}nig et al.~\citep{schoenig.2018} predict next activity values combined with values of the resource attribute and not only next activity values. 
The remaining LSTM architectures \citep{park.2019,di.2017,camargo.2019,weinzierl.2020b,metzger.2019}, except the one of Evermann et al.~\citep{evermann.2017}, are based on the LSTM architecture of~Tax et al.~\citep{tax.2017}.
Among these, Camargo et al.~\citep{camargo.2019} present a context-sensitive extension of Tax et al.~\citep{tax.2017}'s LSTM architecture and report for most of the event logs a higher predictive quality than Tax et al.~\citep{tax.2017} and Evermann et al.~\citep{evermann.2017}.  
Further, the two adaptions are: first, removal of the next timestamp and next resource attribute prediction since we focus on the next activity prediction; second, removal of the embedding layer to be more flexible regarding the attribute encoding.

Moreover, Al-Jebrni et al.~\citep{al.2018} present a DNN architecture with an embedding layer followed by five convolutional blocks for the next activity prediction.
They only feed activity attributes to their DNN architecture. 
Pasquadibisceglie et al.~\citep{pasquadibisceglie.2019} propose a DNN architecture with three convolutional blocks. 
The authors accumulate the onehot encoding of activities in the curse of a process instance (i.e. they count the occurred, onehot encoded activities over time). Additionally, they consider the time difference in days between the current and the first event in a process instance.   
Di Mauro et al.~\citep{dimauro.2019} present a DNN architecture with an embedding layer and three stacked CNN inception modules. They embed activities and consider the time difference between the current and the first event in a process instance.  
We add the CNN architecture of~Al-Jebrni
et al.~\citep{al.2018} to our evaluation since the authors report the highest predictive quality. Again, we remove the embedding layer from the DNN architecture to be more flexible regarding the attribute encoding.   

%
%
%
Last, Khan et al.~\citep{khan.2018} and Lin et al.~\citep{lin.2019} propose encoder-decoder architectures.
Khan et al.~\citep{khan.2018} adapt a differential neural computer~\citep{graves.2016}, an encoder-decoder network with an externalised state memory, for the next activity prediction.
Lin et al.~\citep{lin.2019} present an encoder-decoder network for predicting the next activities as well as assigned attribute values. All categorical attributes are embedded. 
However, encoder-decoder architectures are geared to solve sequence-to-sequence problems 
and not sequence-to-element problems~\citep{goodfellow.2016}.
Therefore, we exclude these architectures from our evaluation. 
%
%
%

\section{Experimental settings}
\label{sec:setting}
\subsection{Description of the event log characteristics}
We use five real-life event logs in our evaluation. Each event log only includes activities executed by humans and not by machines. Such event logs are of particular interest for our evaluation since we focus on the anticipation of future process behaviour. 

\textbf{bpi2013i\footnote{https://data.4tu.nl/repository/uuid:a7ce5c55-03a7-4583-b855-98b86e1a2b07.}:} The event log from the~\textit{BPI challenge 2013} was provided by Volvo IT Belgium. It contains event data of an incident and problem management system. 
The event log is split into sub-logs, namely ``Problems" and ``Incidents". However, we do not consider the ``Problems" event log because it includes a limited number of events. 

\textbf{bpi2017w\footnote{https://data.4tu.nl/repository/uuid:5f3067df-f10b-45da-b98b-86ae4c7a310b.}:} The event log of the \textit{BPI challenge 2017} includes event data about a loan application process from a dutch financial institute. The process considers three types of events: \emph{Application state changes}, \emph{Offer state changes} and \emph{Workflow events}. We only consider workflow events since humans executed these events.
%

\textbf{bpi2019\footnote{https://data.4tu.nl/repository/uuid:a7ce5c55-03a7-4583-b855-98b86e1a2b07.}:} The event log from the \textit{BPI challenge 2019} was provided by a company for paints and coatings. It depicts a purchase order handling process. 
Due to the high computation effort required to apply the entire log, we only consider sequences of $250$ events or shorter and use a 10\%-random-sample from the remaining sequences. 

\textbf{helpdesk\footnote{https://data.mendeley.com/datasets/39bp3vv62t/1.}:} This event log contains data of a ticketing management process form an Italian software company. 

\textbf{fsp:} This event log originates from a financial service provider. It stores the software usage behaviour of the customers. 

\begin{figure}[htb]
\centering
\includegraphics[width=8.0cm]{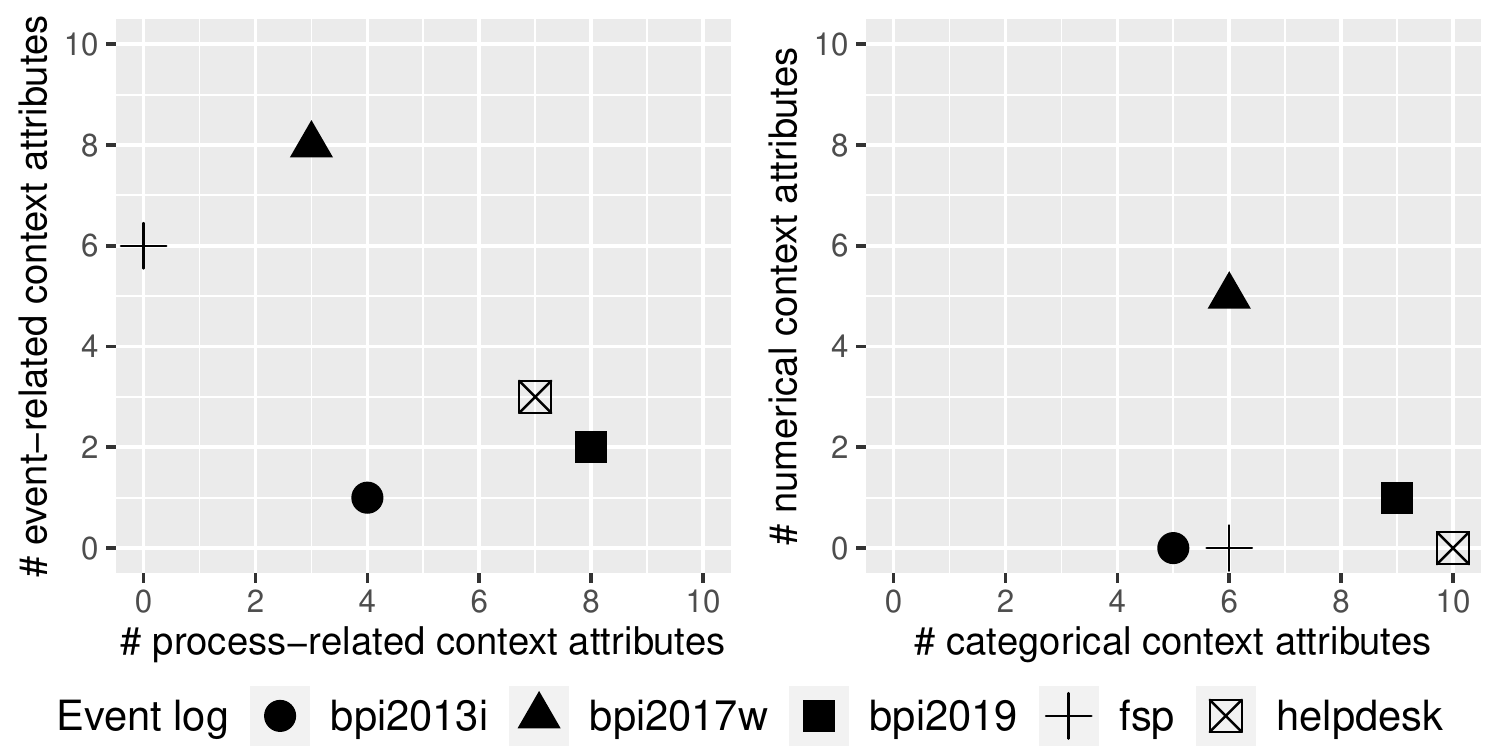}
\vspace{-0.4cm}
\caption{Event logs categorised by scale or type of their context attributes.} 
\label{fig:attribute_characteristics}
\end{figure}

In all event logs, we remove context attributes with more than $600$ values because we assume that these attributes have a limited contribution to a better generalisation\footnote{Please refer to the GitHub project for further details on the pre-processed event logs.}. 
Table \ref{tab: event_logs} shows the event logs' characteristics. Note that we have marked the distinct values of numerical context attributes with * since each attribute value can be different from each other. 
To better understand the event logs' context information, Figure~\ref{fig:attribute_characteristics} categorises these by the scale and type of their context attributes.
%
%

\begin{table}[htb]
\caption{Overview of used real-life event logs.}
\label{tab: event_logs}
\begin{adjustbox}{max width=\textwidth}
\begin{tabular}{|l|l|l|l|l|l|l|}
\hline
\textbf{Id} & \textbf{Characteristics} & \textbf{bpi2013i} & \textbf{bpi2017w} & \textbf{bpi2019} & \textbf{helpdesk} & \textbf{fsp} \\ \hline
(1) & \# instances & 7,553 & 27,373 & 24,938 & 4,580& 2,142   \\
\hline
(2) & \# instance variants & 2,664 & 3,314 & 1,656 & 226 & 1,529 \\
\hline
(3) & (1)/(2)-ratio & 2.84 & 8.26 & 15.06 & 20.27 & 1.40 \\\hline
(4) & \# activity classes &  13 & 8 & 34 & 14 &  161  \\\hline
(5) & \# events& 65,533 & 194,601 & 104,074 & 21,348 &  66,233  \\\hline
(6) & \# events per instance$^{1}$ & [1;123;9;6]& [1;118;8;7] & [1;177;39;27] & [2;15;5;4]  &   [1;247;31;11] \\\hline
(7) & \# activities per instance$^{1}$ & [2;9;6;6]  & [1;6;3;3] & [1;13;7;7] & [2;9;4;4] & [1;34;8;7]   \\\hline
(8) & \# attributes$^{2}$ & [5;0;5] & [11;5;6] & [10;1;9]  & [10;0;10] & [6;0;6] \\\hline
(9) & 
\begin{tabular}[l]{@{}l@{}}\# of an attribute's\\
distinct values\end{tabular} & 
\begin{tabular}[l]{@{}l@{}}[24;25;4;23;32]\end{tabular} & 
\begin{tabular}[l]{@{}l@{}}[2;14;*;*;3;2;\\ 3;514;*;*;*]\end{tabular} & 
\begin{tabular}[l]{@{}l@{}}[20;3;120;6;4;\\ 4;2;2;497;*]\end{tabular} &
\begin{tabular}[l]{@{}l@{}}[22;226;397;21;\\ 7;4;4;4;7;4]\end{tabular} & 
\begin{tabular}[l]{@{}l@{}}[16;44;120;222;\\ 154;46]\end{tabular}\\
\hline
\multicolumn{2}{l}{} $^1$[\text{min; max; mean; median}]$;$ & \multicolumn{2}{c}{} $^2$[\text{total; numerical; categorical}]$;$ &
\multicolumn{2}{l}{} $^*$\text{Numerical attributes.}
\end{tabular}
\end{adjustbox}
\end{table}

%


\subsection{Pre-processing}
A DNN architecture, as used in this paper, requires a tensor-oriented representation for computation, instead of 
raw event log data (log-oriented representation).
Therefore, we transform an event log $\mathcal{L}_{\tau}$ into a data tensor $X$ and a label tensor $Y$ (i.e. next activities). 
Its procedure comprises five steps: 

\textbf{S1: Event log to second-order tensor transformation:} 
%
%
%
%
We transform an event log $\mathcal{L_{\tau}}$ to a second-order tensor $S\in \mathbb{R}^{E \times U}$. 
$E$ is the event log's size $\vert \mathcal{L_{\tau}} \vert$ (i.e. its number of events). $U$ is the event tuple's size $\vert e \vert$ (i.e. an event log's number of attributes). 

\textbf{S2: Missing values handling:} DNNs, as considered in this paper, cannot deal with missing values. 
Therefore, we replace the missing values of numerical attributes by the mean of the existing values; for categorical attributes, we insert the most frequent value of an attribute~\citep{han.2011}.

\textbf{S3: Attribute encoding:} The raw data attributes' representation is not suitable for DNNs, as used in this paper, for two reasons.  
First, its representation obscures valuable information. 
Second, its representation can include textual information that hinders a DNN from calculating forward- and backward propagations.  
%
%
Thus, we encode the activity attribute $f_a(e)$ and each categorical context attribute $d \in $ $f_{cat}(f_{D}(e))$ into a numeric one by applying an encoding technique\footnote{While the encoding techniques ordinal, binary, onehot and hash require as input a vector (all attribute values), word2vec requires as input a set of vectors (each vector is a sequence of attribute values corresponding to a process instance).}. Note, in Section~\ref{sec:encoding}, we briefly describe the encoding techniques we use in our evaluation.
For numerical attributes, we perform min-max normalisation (with $min=0$ and $max=1$) to grand equality among the attributes~\citep{han.2011}.

\textbf{S4: Prefix and label creation:} We create for each $\sigma_c \in S$ with $1 \leq i \leq \vert \sigma_{c} \vert$ a set of prefixes $R$ by applying $hd^{k}(o_{c})$ and a set of next activity labels $K$ via $next(hd^{k}(o_{c}))$ with $0<k<n$.  
We do not create prefixes of size $0$ because it seems unrealistic to predict next activities based on no historical information. 
%

\textbf{S5: Third-order tensor creation:}
Based on the prefix set $R$, we create the third-order tensor $X \in \mathbb{R}^{E \times M \times U}$. 
$M$ is the event log's longest process instance $\vert max_{\sigma}(\mathcal{L_{\tau}}) \vert$.
The remaining space for a sequence $ \sigma_{c}\in X$ is padded with zeros, if $\vert \sigma_{c} \vert < \vert max_{\sigma}(\mathcal{L_{\tau}}) \vert$.
%
%
Concerning $U$, we represent an event $e$ by 
a set of activity attributes, 
a time attribute ($\Delta t_{i}$ = $f_{t}(\sigma_{c}(i))-f_{t}(\sigma_{c}(i-1)$; if $i = 0$, then $\Delta t_{i}=0)$ and
a set of context attributes.
The number of activity and categorical context attributes varies depending on the encoding technique. 
%
%
After constructing $X$, we create a third-order label tensor $Y \in \mathbb{R}^{E \times M \times V}$ based on the label set $K$. 
%
%
$V$ is the number of attributes. We onehot encode the label. Thus, each attribute refers to a next activity class.    
%

\subsection{Deep-neural-network architectures}
%

\begin{figure}[htb]
\centering
\includegraphics[height=2.9cm]{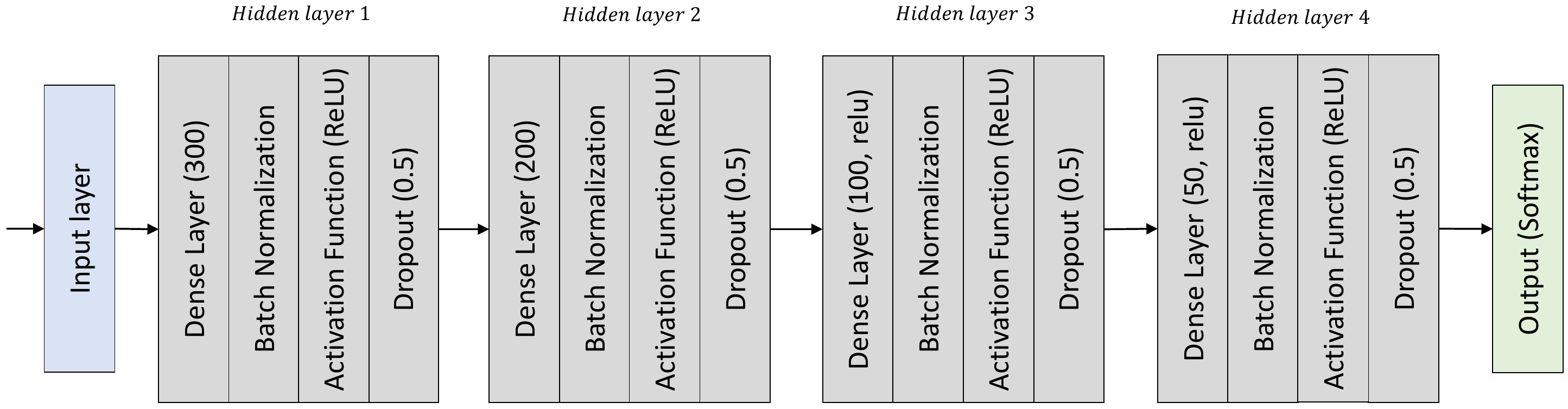}
\vspace{-0.5cm}
\caption{MLP adapted from~Theis and Darabi~\citep{theis.2019}.} 
\label{fig:mlp}
\end{figure}

\textbf{Multi-layer perceptron (MLP):} The MLP architecture in Theis and Darabi~\citep{theis.2019} consists of one input layer (including flattening), four hidden layers and one output layer, as shown in Figure~\ref{fig:mlp}. The hidden layers of input sizes $\{300,200, 100, 50\}$ each have a batch normalisation operator to speed up training~\citep{ioffe.2015} and a \textit{ReLU} activation function followed by random dropout of 50\% of the input units along with their connections~\citep{srivastava.2014}. The output layer consists of a \textit{Softmax} activation function. Each of the dense layers represents a fully connected layer to its input neurons.

\begin{figure}[htb]
\centering
\includegraphics[height=2.9cm]{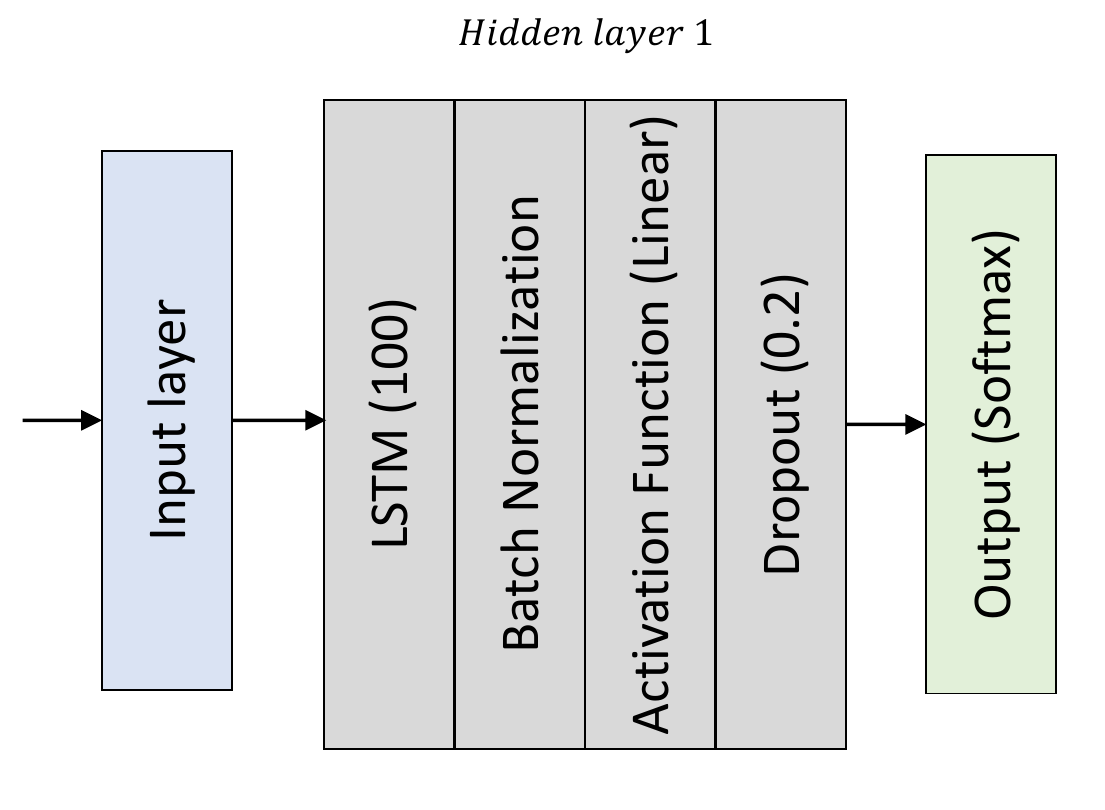}
\vspace{-0.5cm}
\caption{LSTM adapted from~Camargo et al.~\citep{camargo.2019}.} 
\label{fig:lstm}
\end{figure}

\textbf{Long short-term memory neural network (LSTM):} Our adapted version of the LSTM architecture of~Camargo et al.~\citep{camargo.2019} consists of an input layer, one hidden layer and one output layer (cf. Figure~\ref{fig:lstm}).  
The hidden layer represents an LSTM layer with output size 100 followed by a batch normalisation to seed up training, a linear activation function 
and a random dropout of 20\% of the input units together with their connections~\citep{srivastava.2014}.
An LSTM cell~\citep{hochreiter.1997} is a neuron with a memory and gates that control the memory to grasp temporal dependencies in sequential data~\citep{lecun.2015}.
%
%
%
%
%
%

\begin{figure}[htb]
\centering
\includegraphics[height=2.9cm]{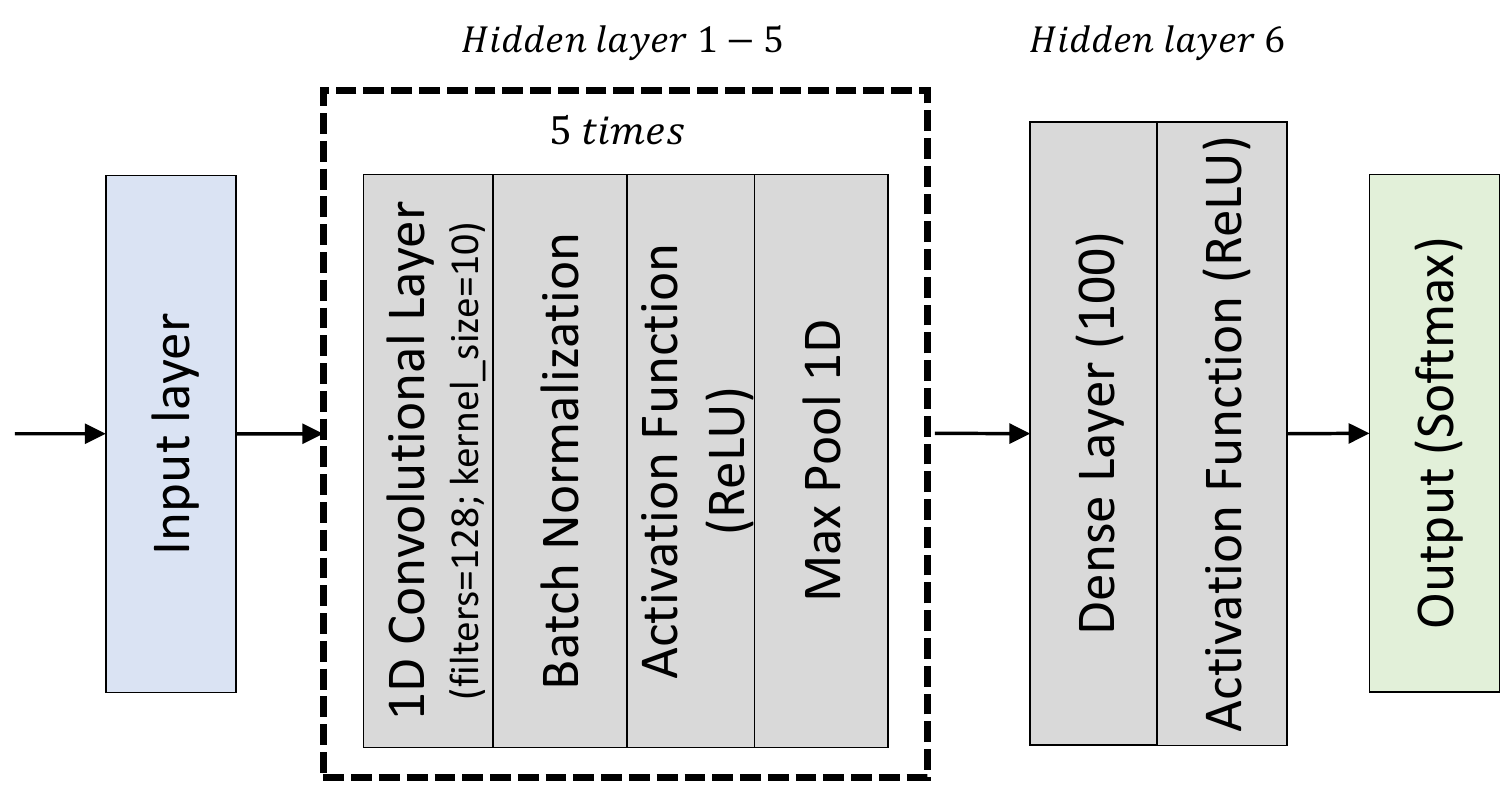}
\vspace{-0.5cm}
\caption{CNN adapted from Al-Jebrni et al.~\citep{al.2018}.} 
\label{fig:cnn}
\end{figure}

\textbf{Convolutional neural network (CNN):} The CNN architecture of~Al-Jebrni et al.~\citep{al.2018} consists of one input layer, seven hidden layers and one output layer, as depicted in Figure~\ref{fig:cnn}.
The first five hidden layers are convolutional blocks.
Each block starts with a one-dimensional convolutional layer to filter more abstract features (feature maps) followed by a batch normalisation~\citep{ioffe.2015} to speed up the training. The output of the batch normalisation runs through the nonlinear activation function \textit{ReLU}. Each block ends with the application of the pooling function \textit{one-dimensional max-pooling} to perform a dimensionality reduction on the time axis~\citep{lecun.2015}.
%
%
%
%
The sixth hidden layer is a dense layer consisting of $100$ neurons and applies the activation function \textit{ReLU}. 
%

\subsection{Parameter learning of the deep-neural-network architectures}
%
Gradient-based optimisation algorithms are applied to update the DNN architectures' parameters, i.e. weights and biases, during training.
%
%
Table~\ref{tab:learning_parameters} summarises the reported hyperparameters determining the DNN architectures' learning procedure. 

\begin{table}[htb]
\centering
\caption{Reported hyperparameters determining learning.}
\label{tab:learning_parameters}
\resizebox{0,6\textwidth}{!}{
\begin{tabular}{|l|c|l|l|}
\hline
\begin{tabular}[c]{@{}l@{}}\textbf{Parameter}\end{tabular} &  
\begin{tabular}[c]{@{}l@{}}\textbf{MLP} \citep{theis.2019}\end{tabular}              & 
\begin{tabular}[c]{@{}l@{}}\textbf{LSTM} \citep{camargo.2019}\end{tabular} & 
\begin{tabular}[c]{@{}l@{}}\textbf{CNN} \citep{al.2018}\end{tabular} 
\\ \hline
Loss function        & \multicolumn{3}{c|}{Categorical cross-entropy}                                \\ \hline
Optimiser          & \multicolumn{1}{l|}{Adam}  & Nadam         & Adam  \\      
\hline
Learning rate      & \multicolumn{1}{l|}{0.001} & 0.002         & 0.001   \\     \hline

\end{tabular}
}
\end{table}

Theis and Darabi~\citep{theis.2019} and Al-Jebrni et al.~\citep{al.2018} perform \textit{Adam} as optimisation algorithm for the MLP respective the CNN with a learning rate of $0.001$. 
Camargo et al.~\citep{camargo.2019} use \textit{Nadam} for the \textit{LSTM} with a learning rate of $0.002$. 
All works calculate the loss based on the \textit{categorical cross-entropy} loss function.
For other hyperparameters of the optimisation algorithms, if not specified in the original works, we use the default values.
In line with Al-Jebrni et al.~\citep{al.2018}, Camargo et al.~\citep{camargo.2019} and Theis and Darabi~\citep{theis.2019}, we set the number of epochs to $100$.
To speed up model learning, we apply a batch normalisation with a batch size of 128, where gradients are updated after each 128\textsuperscript{th} instance of the training set. 
We set the batch size to $128$ since bigger sizes tend to sharp minima and, generally, a sharp minimum leads to a more reduced generalisation~\citep{keskar.2016}.     

\subsection{(Categorical) encoding techniques}
\label{sec:encoding}
\begin{figure}[htb]
\centering
\includegraphics[width=8cm]{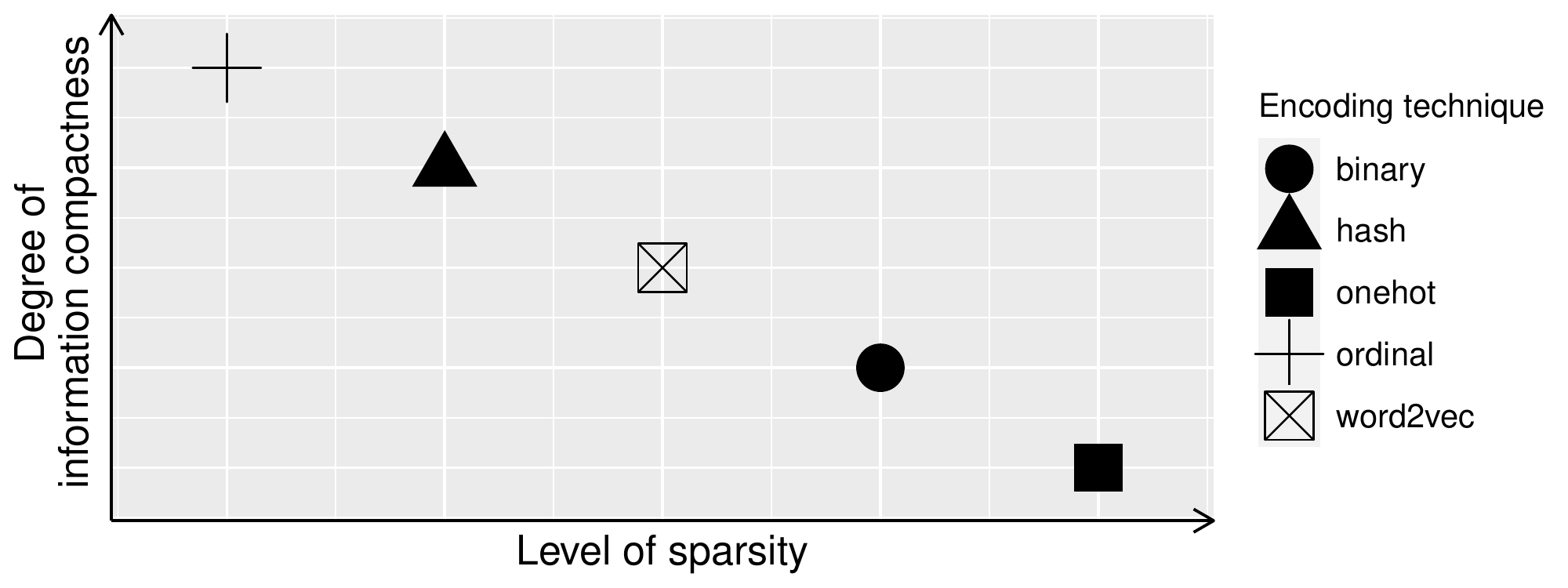}
\vspace{-0.7cm}
\caption{Categorisation of the encoding techniques.} 
\label{fig:encoding_techniques}
\end{figure}

Concerning data encoding, PBPM researchers differ between trace abstraction techniques and feature extraction functions~\citep{teinemaa.2019}.  
We encode sequences with the trace abstraction technique~\textit{index-based encoding} because it considers complete prefixes, and therefore avoids information loss~\citep{leontjeva.2016}.
%
%
From each categorical attribute of an \textit{index-based} encoded sequence, a more meaningful attribute or set of attributes for ML algorithms can be extracted by using a feature extraction function. In the context of ML, these functions are called encoding techniques~\citep{potdar.2017}. 
We include five encoding techniques with different characteristics to our evaluation (cf. Figure~\ref{fig:encoding_techniques}). 
Each technique is categorised by the dimensions \textit{degree of information compactness} and \textit{level of sparsity} (i.e. number of extracted attributes).
%
%
%

\textbf{Ordinal encoding:} Each value of a categorical attribute is mapped to an integer value. 
It does not add any new attributes to the data (low level of sparsity), and implies an order that may not exist (high degree of information compactness)~\citep{voneye.1996}. 

\textbf{Hash encoding:} Each value of a categorical attribute is mapped through a hash function to a hash vector with a fixed size (low to intermediate level of sparsity)~\citep{weinberger.2009}. 
Some loss of information may exit due to collisions (intermediate to high degree of information compactness). 
According to~Mehdiyev et al.~\citep{mehdiyev.2018}, we set the dimensionality of an attribute's hash vectors to $10$. Finally, we use the default hash function~\textit{md5}.
    
\textbf{Word2vec encoding:} Each value (``word") of a categorical attribute is represented by an embedding vector with a fixed size (intermediate level of sparsity). The vector's dimensionality refers to the number of neurons of a shallow neural network's hidden layer and its values to the neurons' weights. Since the neural network considers the context of attributes' values during training, the resulting embedding vectors express the similarity of attribute values (intermediate level degree of information compactness).  
To create the embedding vectors, we apply the approach \textit{common bag of words (CBOW)}~of~Mikolov et al.~\citep{mikolov.2013}.
We set the embedding vectors' dimensionality to $32$, 
initial learning rate to $0.025$ (with a decrease per epoch by $0.002$), 
window size to $5$ and 
number of epochs to $10$. 
For other hyperparameters, we use the default values.  

\textbf{Binary encoding:} Each value of a categorical attribute is mapped to binary digits (low to intermediate degree of information compactness). 
It creates for every binary digit a new attribute to the data (intermediate to high level of sparsity) \citep{potdar.2017}.

\textbf{Onehot encoding:} Each value of a categorical attribute becomes a new attribute (high level of sparsity) with possible values $1$ and $0$ denoting the presence or absence of the specific value (low degree of information compactness)~\citep{potdar.2017}. 

\subsection{Validation procedure}
For each experiment (i.e. a DNN architecture combined with an encoding technique), we perform ten-fold cross-validation with random instance-based sampling to ensure model generalisation~\citep{kohavi.1995}. 
Thereby, we consider 90\% of an event log's process instances for training and 10\% for testing.
Further, we use 10\% of the training set for validation. 
We train the models based on the remaining training set. For testing, we select the best model by applying early stopping~\citep{goodfellow.2016} based on the validation set. 
If the validation loss does not decrease over ten epochs, the model with the lowest validation loss is used, and the training procedure is stopped.

%
%
Before splitting the event log, we randomly shuffle the process instances (instance-based sampling) to improve model generalisation further.
We prefer instance-based sampling over event-based sampling~\citep{tama.2019}, since it does not ignore the instance-affiliation of event log entries.
%
In the next activity prediction, instance-based sampling represents a more realistic approach because it leaves the process instances intact. 

To measure predictive quality, we calculate three metrics: average weighted 
\emph{Accuracy},
\emph{F1-Score} and
$AUC_{PR}$.
%
These ML metrics are well-established in the PBPM community and, for instance, a further explanation can be found in Mehdiyev et al.~\citep{mehdiyev.2018}.
%
%
%
\textit{Accuracy} is the most commonly used metric in the PBPM domain~\citep{marquez.2017a} and represents the overall correctness of a model.
%
\textit{F1-Score} is the harmonic mean of \emph{Precision} and \textit{Recall}. 
%
%
%
%
%
%
$AUC_{PR}$ is the area under the precision-recall curve. It is suitable for data with a strongly imbalanced class distribution~\citep{davis.2006}, as can be the case with activities in event logs.  
%
%
%
%
For all metrics, we calculate the \textit{mean} and the \textit{standard deviation} over all ten folds of an experiment.
%
In our experiments, we do not optimise the hyperparameters of the various models, but derived them from previous research in the literature, because of the high number of experiments in our evaluation.

\subsection{Implementation}
We conducted all experiments on a workstation with 12 CPU cores, 128 GB RAM and a single GPU NVIDIA Quadro RXT6000. We implemented the experiments in \textit{Python} 3.7. We used the DL library \textit{Keras}\footnote{https://keras.io.} 2.2.4 with the \textit{TensorFlow}\footnote{https://www.tensorflow.org.} 1.14.1 backend
to build DL models. We implemented word2vec based on \textit{Gensim}\footnote{https://radimrehurek.com/gensim.}
and other categorical-encoding techniques with \textit{categorial-encoding}\footnote{https://github.com/scikit-learn-contrib/categorical-encoding.}.
Finally, the source code is available on GitHub\footnote{https://github.com/fau-is/nap-dnn-c.}.

\section{Experiment results}
\label{sec:results}
\begin{figure}[!ht]
\begin{subfigure}
\centering
\includegraphics[width=\linewidth]{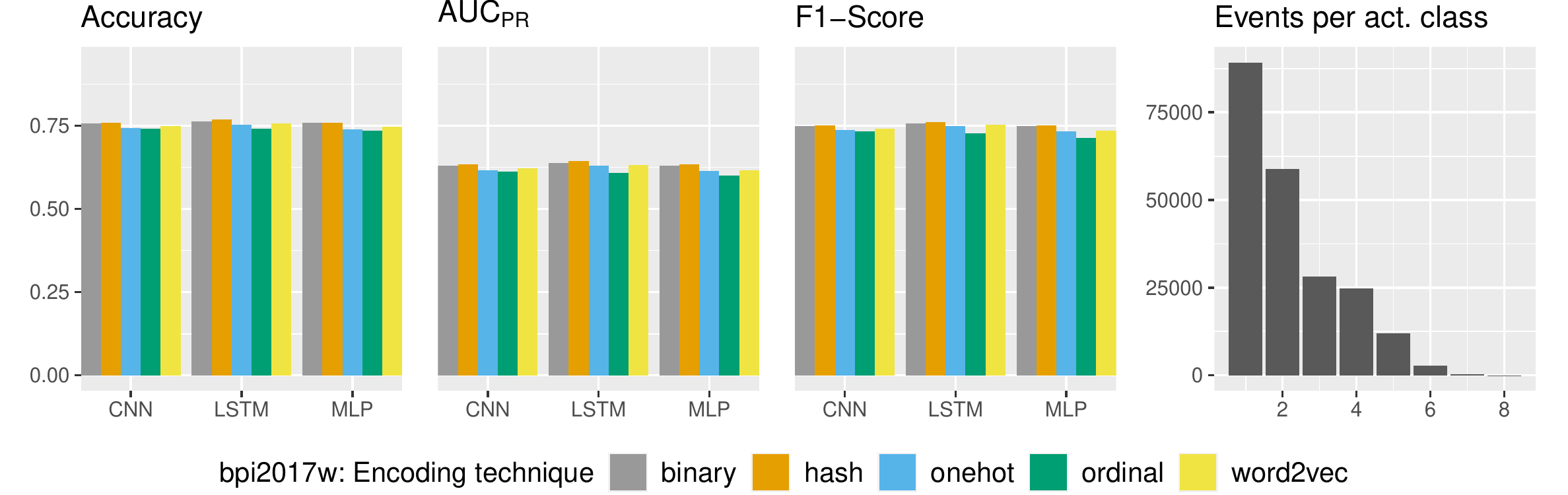}
\vspace{-1cm}
\caption{Benchmark metrics for the \textit{bpi2017} event log.}
\label{fig:metrics_bpi2017}
\end{subfigure}

\vspace{0.3cm}

\begin{subfigure}
\centering
\includegraphics[width=\linewidth]{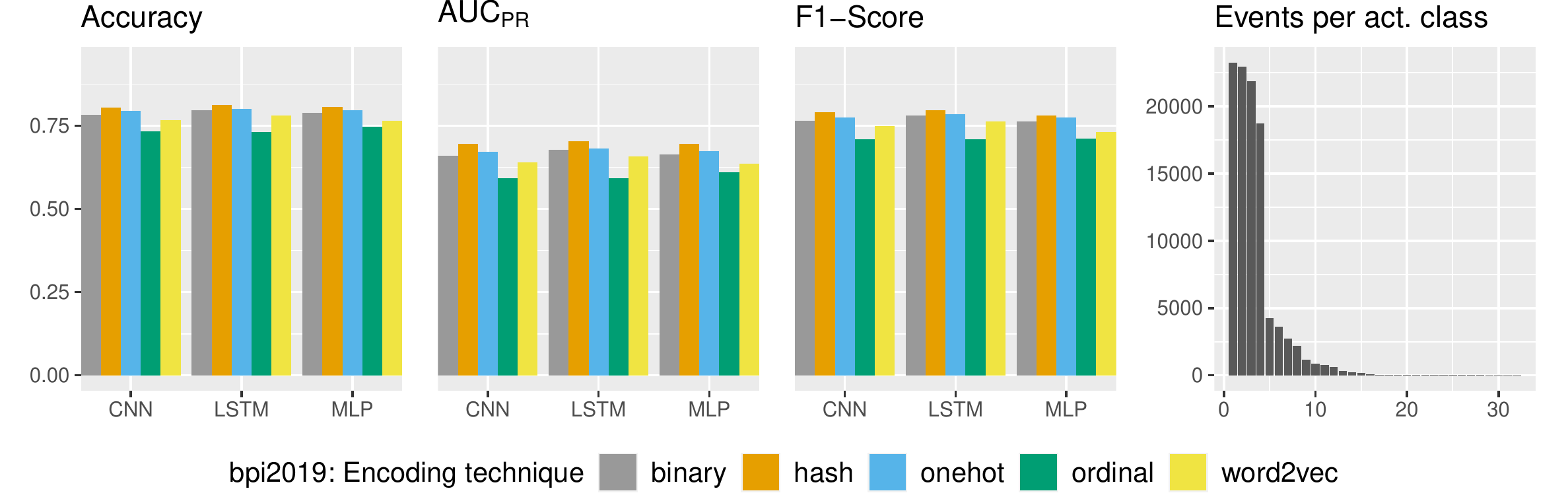}
\vspace{-1cm}
\caption{Benchmark metrics for the \textit{bpi2019} event log.}
\label{fig:metrics_bpi2019}
\end{subfigure}

\vspace{0.3cm}

\begin{subfigure}
\centering
\includegraphics[width=\linewidth]{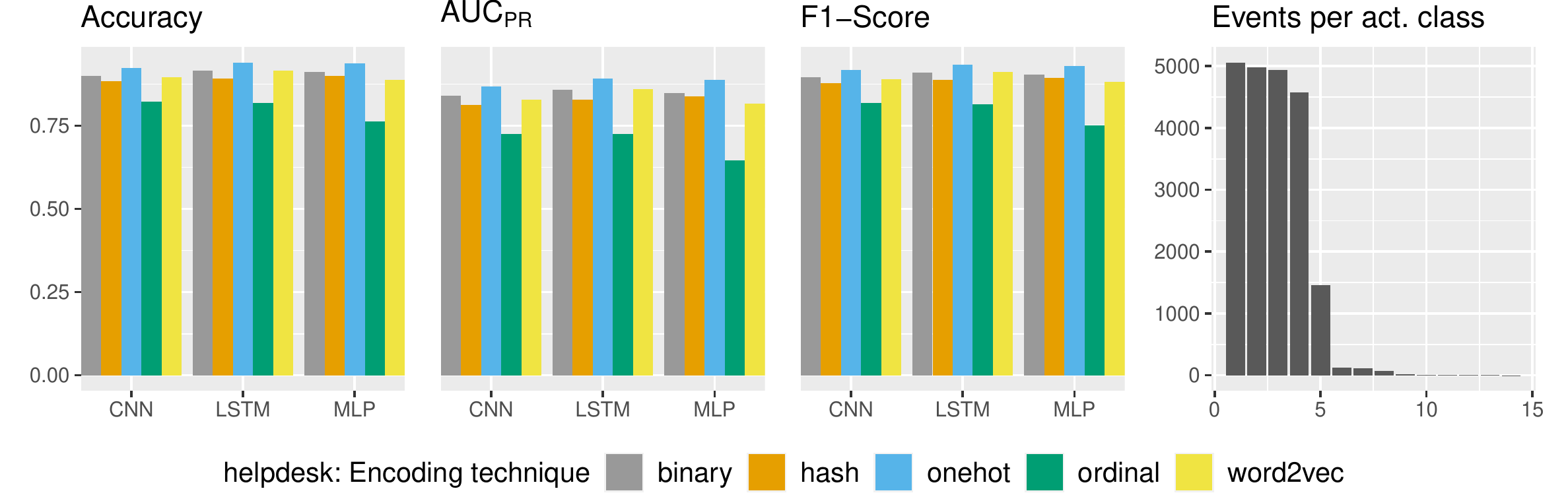}
\vspace{-1cm}
\caption{Benchmark metrics for the \textit{helpdesk} event log.}
\label{fig:metrics_helpdesk}
\end{subfigure}
\end{figure}

\begin{figure}[h!]
\begin{subfigure}
\centering
\includegraphics[width=\linewidth]{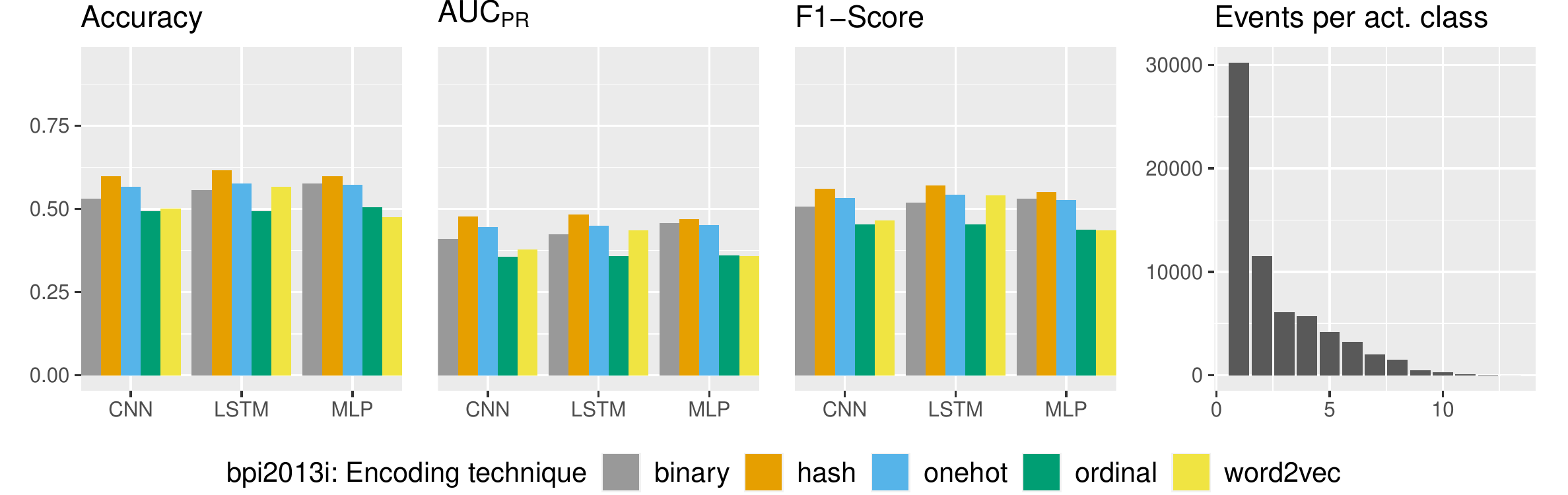}
\vspace{-1cm}
\caption{Benchmark metrics for the \textit{bpi2013i} event log.}
\label{fig:metrics_bpi2013}
\end{subfigure}

\vspace{0.3cm}

\begin{subfigure}
\centering
\includegraphics[width=\linewidth]{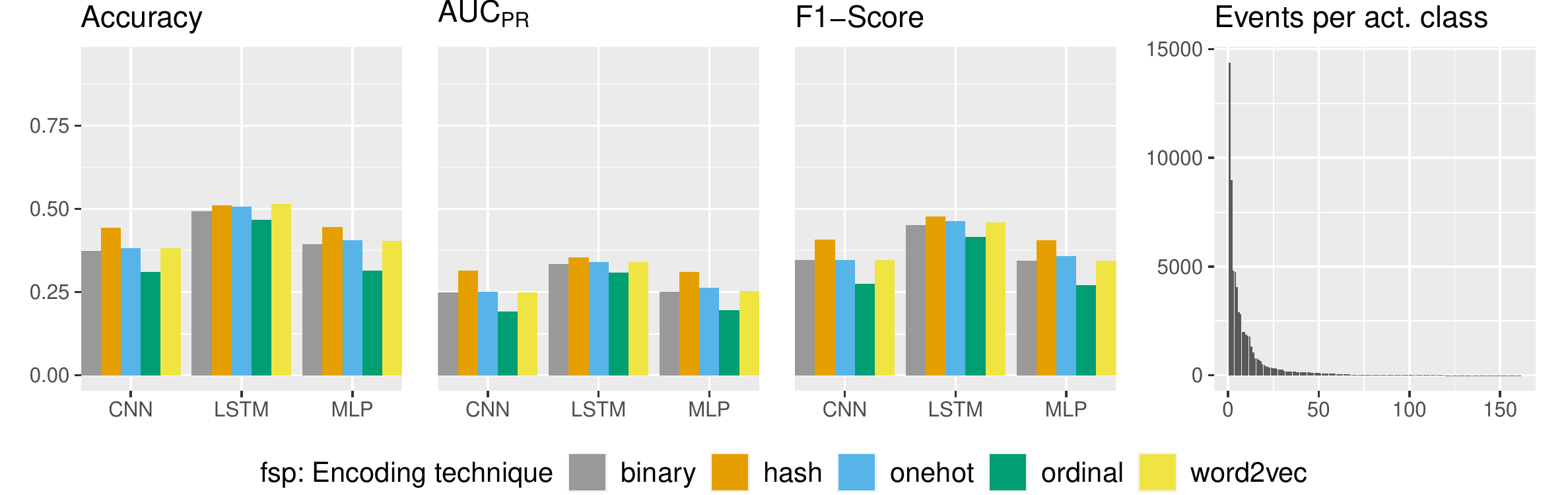}
\vspace{-1cm}
\caption{Benchmark metrics for the \textit{fsp} event log.}
\label{fig:metrics_fsp}
\end{subfigure}

\end{figure}

%
%
We report the \textit{Accuracy}, \textit{F1-Score} and \textit{$AUC_{PR}$} for each event log by DNN architecture and encoding technique. 
Figures~\ref{fig:metrics_bpi2017} - \ref{fig:metrics_fsp} present the calculated mean results graphically per event log. In each of the afore-mentioned figures, the x-axis represents the three DNN architectures (MLP, LSTM and CNN). The differently coloured lines show the values of the particular metric, which was achieved with one of the five encoding techniques (binary, hash, ordinal, onehot and word2vec). A detailed result table for all metrics, which also includes the standard deviation of the ten folds, can be found in the Appendix (cf. Table \ref{fig:accuracy}). Analysing the results, we were able to identify a pattern of behaviour among the event logs. Thus, we will present the results in two groups. 

The first group of results shows the benchmark for the event logs \textit{bpi2019, bpi2017w} and \textit{helpdesk} (cf. Figures~\ref{fig:metrics_bpi2017}, \ref{fig:metrics_bpi2019} and \ref{fig:metrics_helpdesk}). Here, \textit{bpi2019} shows a ranking of encoding techniques throughout all three metrics with hash, onehot, binary and word2vec at the top and ordinal at the bottom end. 
Within the top four encoding techniques, there are only small differences, and the choice of DNN architecture does not seem to have a considerable effect either.  
Note that while binary, hash, onehot and word2vec perform very similar for \textit{bpi2017w} and \textit{bpi2019}, onehot outperforms the other three by $0.04$ (average of all metrics) for the \textit{helpdesk}. Overall, the \textit{helpdesk} shows the highest metric values, followed by \textit{bpi2019} and \textit{bpi2017w}. The better performing encodings achieve an average \textit{Accuracy} of 0.79 - 0.75 - 0.91 (\textit{bpi2019} - \textit{bpi2017w} - \textit{helpdesk}), an average $AUC_{PR}$ of 0.67 - 0.63 - 0.85 and an average \textit{F1-Score} of 0.77 - 0.75 - 0.90.

The fourth chart of each event log reports the number of events (y-axis) per activity class (x-axis) of the whole event log. Figures~\ref{fig:metrics_bpi2017}, \ref{fig:metrics_bpi2019} and \ref{fig:metrics_helpdesk} show that the number of activity classes for those event logs range from $8$ to $34$ (cf. Table \ref{tab: event_logs}). 
All event logs have class imbalances in terms of activities. \textit{Bpi2019} and \textit{helpdesk} both have four majority classes, which have similar numbers of events, and several minority classes. \textit{Bpi2017w} has the lowest quantity of overall classes with a less abrupt decrease in numbers of events in comparison to the other two logs. But also here we find multiple majority classes.

The second group of results, which includes the \textit{bpi2013i} and \textit{fsp} event logs, shows different patterns (cf. Figures \ref{fig:metrics_bpi2013} and \ref{fig:metrics_fsp}).
Even though there also exists a ranking of encoding techniques throughout all DNN architectures and metrics, the spread between the different techniques is much more equally distributed. No certain cluster emerges as in the previous group. Here, the hash encoding leads in front of onehot, binary, word2vec and ordinal. 

Additionally, the LSTM shows a peak and an improvement in comparison to all other architectures , which is most prevalent in the \textit{fsp} event log (+0.09 in average over all metrics), for all encoding techniques. 
Overall, the metric values for the second group are worse than for the first group. The average \textit{Accuracy} values for \textit{bpi2013i} are 0.55 - 0.60 - 0.50 - 0.57 - 0.51 (binary - hash - ordinal - onehot - word2vec). \textit{$AUC_{PR}$} averages at 0.43 - 0.48 - 0.36 - 0.45 - 0.39 and the \textit{F1-Score} at 0.52 - 0.56 - 0.45 - 0.53 - 0.48. The \textit{fsp} event log performs even worse with \textit{Accuracy} values between 0.31 and 0.52, \textit{$AUC_{PR}$} between 0.19 and 0.35 and \textit{F1-Score} between 0.27 and 0.48.
Both event logs also show a strong class imbalance regarding the activity classes. \textit{Bpi2013i} consists of $13$ classes and the \textit{fps} event log of $161$ classes. However, the figures show that (contrary to the first group) both event logs have only one prevalent majority class and a high number of minority classes.


\section{Discussion}
\label{sec:discussion}

\subsection{Findings}
\textbf{F1: The encoding technique has a stronger impact on the predictive quality of DNNs than the type of architecture.}
In both result groups, we report that a ranking of encoding techniques persists throughout the DNN architectures in all three metrics. 
Based on this observation, we took a closer look into the underlying data. 
Table \ref{tab:avg_techniques_architectures} shows the average values of the ten folds of all encoding techniques and DNN architectures over all event logs per metric. In addition, we included the average over the three metrics and in each case the deviation from the average of all rows. 
We sorted the table in descending order based on the average values of all metrics. 
The average values of the DNN architectures and encoding techniques (rows) show how well the respective encoding technique/DNN architecture performs over all metrics and event logs. 
It is an indicator of the impact of DNN architecture and/or encoding technique choice on the predictive quality when disregarding other event log properties. Aside from the previously identified ranking, the deviation values indicate that the choice of the encoding technique has a stronger impact on the predictive quality than the DNN architecture. Except for LSTM, all the encoding techniques are situated at the very top or bottom of the ordered table, which means a stronger impact, while the MLP and CNN architectures can be found in the middle. The average of the deviations of the encoding techniques is higher compared to the DNN architectures' average.
A stronger impact does not equal a more positive effect. However, Table \ref{tab:avg_techniques_architectures} also shows that the better performing encoding techniques achieve better predictive quality with 1.54\% to 4.72\% higher average metrics. The simple summation of different quality metrics, which evaluate different aspects and are based on different data points, is generally not a good practice. 
However, in this case, it serves the purpose of comparing the multitude of results. Additionally, identical patterns, with similar deviation values, are also present for each metric, as Table \ref{tab:avg_techniques_architectures} shows.

\begin{table}[htb]
\caption{Averages of encoding techniques and architectures over all event logs.}
\begin{threeparttable}
\label{tab:avg_techniques_architectures}
\vspace{-0.5cm}
\centering
\resizebox{0.8\textwidth}{!}{%
\begin{tabular}{|p{2.2cm}|l|l|l|l|l|l|l|l|}
\hline
\textbf{Arch./Tech.} & \multicolumn{2}{l|}{\textbf{Accuracy$^*$}} & \multicolumn{2}{l|}{\textbf{$AUC_{PR}^*$}} & \multicolumn{2}{l|}{\textbf{F1-Score$^*$}} & \multicolumn{2}{l|}{\textbf{Avg. of metrics$^*$}} \\ \hline
Hash                              & 0.7061               & (+3.95\%)      & 0.5928              & (+5.86\%)      & 0.6839               & (+4.54\%)      & 0.6609              & (+4.72\%)      \\
\hline
LSTM                              & 0.6995               & (+2.97\%)      & 0.5781              & (+3.23\%)      & 0.6771              & (+3.50\%)      & 0.6515              & (+3.23\%)      \\
\hline
Onehot                            & 0.6960               & (+2.46\%)      & 0.5822              & (+3.96\%)      & 0.6736               & (+2.97\%)      & 0.6506              & (+3.09\%)      \\
\hline
Binary                            & 0.6865               & (+1.06\%)      & 0.5712              & (+2.00\%)      & 0.6646               & (+1.59\%)      & 0.6408              & (+1.54\%)      \\
\hline
Word2vec                          & 0.6797               & (+0.00\%)      & 0.5548              & (-0.93\%)      & 0.6495               & (-0.72\%)      & 0.6280              & (-0.49\%)      \\
\hline
MLP                               & 0.6728               & (-0.96\%)      & 0.5511              & (-1.59\%)      & 0.6402               & (-2.14\%)      & 0.6213              & (-1.55\%)      \\
\hline
CNN                               & 0.6656               & (-2.02\%)      & 0.5507              & (-1.66\%)      & 0.6454               & (-1.35\%)      & 0.6206              & (-1.66\%)      \\


\hline
Ordinal                           & 0.6280               & (-7.55\%)      & 0.4987              & (-10.95\%)      & 0.5996               & (-8.35\%)      & 0.5755              & (-8.81\%)      \\
\hline
\textbf{Average}                  & \textbf{0.6793}      &                & \textbf{0.5600}     &                & \textbf{0.6542}      &                & \textbf{0.6311}     &           
\\    
\hline
\end{tabular}
}
\begin{tablenotes}
\scriptsize
\item[$*$] Values in brackets represent the percentage deviation from the column-average of the metric.
\end{tablenotes}
\end{threeparttable}
\end{table}

\textbf{F2: The hash encoding achieves the highest predictive quality.}
Second, we can show that the hash encoding achieves the highest predictive quality in regard to \textit{Accuracy}, \textit{$AUC_{PR}$} and \textit{F1-Score} based on the average of all event logs (cf. Tables \ref{tab:avg_techniques_architectures} and \ref{fig:accuracy}). In addition, this is valid regardless of the DNN architecture. %
On an event-log level, the hash encoding performs best for the four tested event logs \textit{bpi2013i}, \textit{bpi2017w}, \textit{bpi2019} and \textit{fsp}. Note that for the \textit{fsp} the word2vec encoding could achieve a higher \textit{Accuracy} for the LSTM, however, it is still fairly close to the hash encoding and additionally we can show that the hash encoding performs best for both the \textit{$AUC_{PR}$} and \textit{F1-Score}. 
Solely for the \textit{helpdesk} event log the onehot encoding results in better predictive quality in regards to the evaluated metrics over all DNN architectures.
To further evaluate this observation, we conduct a significance test to check whether the metrics are significantly different from each other. 
First, we use the \textit{Friedman test} \citep{friedman.1940} to analyse whether the results show a significant difference. Second, we apply the \textit{Nemenyi test} \citep{nemenyi.1962}\footnote{Please refer to the GitHub project for the results.}. It performs a pair-by-pair comparison to find out which of the experiments are significantly different from one another.
Comparing the hash and onehot encoding for the \textit{helpdesk} event log, we cannot observe a significant difference in the predictive quality for each proposed DNN architecture.
For the \textit{fsp} event log, the \textit{Accuracy} of the LSTM is higher with the word2vec than with the hash encoding. However, the values are not significantly different. 
For the $AUC_{PR}$ and the \emph{F1-Score}, the results are better with the hash encoding.
Over all event logs, the word2vec performs worse in regards to predictive quality than the hash or onehot encoding regardless of the DNN architecture.
One reason for the good predictive quality of the hash encoding could be the attribute dimensions. Hash encoding can  deal better with attributes with a wider variety of values, such as for the \textit{fsp} event log. The onehot encoding reports best results for attributes with a small variety of possible values, e.g. the \textit{helpdesk} event log. 
For the ordinal encoding, we receive on average the worst results regardless of the encoding technique for each metric. Since an integer value is assigned to each category of an attribute, an order is created, which might not exist in the raw data \citep{voneye.1996}. Therefore, the ordinal encoding can be biased to a certain degree, which influences the learning procedure of a DNN negatively. 
The same problem applies to binary encoding, where values of each attribute are translated to a binary string~\citep{potdar.2017}. Thus, the values might be falsely assigned a distinct order. 

\textbf{F3: The LSTM achieves the highest predictive quality.}
Third, we can show that the LSTM achieves higher predictive quality than the other proposed DNN architectures in regards to \textit{Accuracy}, \textit{$AUC_{PR}$} and \textit{F1-Score} based on the average of all event logs (cf. Tables \ref{tab:avg_techniques_architectures} and  \ref{fig:accuracy}). Besides, this is valid regardless of the encoding technique.
On an event-log level the overall highest predictive quality is always achieved with the LSTM.
%
%
Looking at the values for each encoding technique, the LSTM outperforms the other applied DNN architectures in the \textit{fsp} for all evaluated encoding techniques regarding predictive quality. 
%
For the other event logs, there are some exceptions mainly in favour of the MLP. 
For instance, given the binary encoding, the \textit{Accuracy} for the \textit{bpi2013i} is, absolutely seen, higher with an MLP than with the LSTM. 
However, for all those cases, our significance analysis shows that these exceptions are not significantly different from the one with the LSTM.
%

\textbf{F4: The magnitude of class imbalance and instance to variants ratio impacts the overall predictive quality.}
All event logs present unbalanced data, such that the frequency of occurrence of the different event types is unequally distributed. 
The event logs of the second group, \textit{fsp} and \textit{bpi2013i}, have one prevalent majority class and various minority classes, while the event logs of group one present up to four majority classes. 
This shows that while both groups are prone to class imbalance, the magnitude of the class imbalance is different. 
This could be an indicator of the overall worse predictive quality of the second group. In this regard, the \textit{$AUC_{PR}$} metric, which reports a specifically poor predictive quality in the second group, is of particular interest, since it is a more informative metric under data with highly imbalanced classes. 
Another important value to consider, in comparing the two groups of results, is the instance/variants ratio of the underlying event logs (cf. Table \ref{tab: event_logs}). 
It examines the relation between the number of variants (different paths through the business process) and the number of instances (total executions of the business process). 
The event logs of the first group, for which a higher predictive quality can be achieved, have a greater instance/variants ratio of 8.26 up to 20.27 instances per variant. 
In contrast, the event logs of the second group only have 1.40 and 2.84 instances per variant. 
Alongside with the very high amount of activity classes in the \textit{fsp} event log, this could be one reason for the difference in the overall predictive quality of the two groups. 
%

\subsection{Limitations}
Even though our work provides new insights regarding the interplay of DNN architectures and the attributes' encoding for the next activity prediction, it includes two shortcomings.
First, we did not perform a hyperparameter optimisation due to the high number of experiments.  
Instead, for every evaluated DNN architecture, we took over the hyperparameter values from the original papers. 
%
%
Second, we might not have included all existing DNN architectures for next activity prediction.
This is attributed to the fact that DL is gaining momentum in the BPM community and a lot of researchers are working on novel approaches for solving PBPM tasks, such as the next activity prediction.
%
%

\subsection{Future research}
We identified five main directions for future research.
First, researchers can build on our work and conduct further experiments based on other real-life event logs to gain a better understanding of the relationship between a process's data representation and the DNN architectures' quality for next activity predictions. A comprehensive hyperparameter optimisation, e.g. through random search~\citep{bergstra.2012}, could provide new insights. 
%
Another avenue for future research is to investigate new forms of DNN architectures for the next activity prediction. 
In this course, we call for a transfer and an adaption of proven DNN architectures from the broader field of time series classification~\citep{fawaz.2019} to the PBPM domain. 
%
Further, novel approaches for representing an event log's instances can further improve the predictive quality of DNNs. 
For instance, concurrency-based
approaches~\citep{hinkka.2019,evermann.2017} seem promising, where information of parallel executed instances is included in an attribute vector.
Another example could be a characteristic-based approach, in which each context attribute of an event log is encoded with a different technique depending on characteristics like the number of values or the number of missing values.
Researchers can also investigate the effect of class imbalance on the quality of next activity predictions which is triggered by the "happy path" phenomenon in process data. 
The ML literature is helpful here since it provides many approaches to address the imbalanced class problem~\citep{he.2009}.  
%
Another direction of future research is to reduce the decision space for next activity predictions. Even though predictions are only necessary at decision points, only very view works like~Metzger et al.~\citep{metzger.2019} learn DNNs in a targeted manner.
%

\section{Conclusion}
\label{sec:conclusion}
A plethora of DL-based PBPM techniques for the next activity prediction have been proposed by researchers. 
The proposed techniques strongly vary in terms of the type and structure of DNN architecture and the encoding of the process instances' attributes. 
%
Therefore, it is challenging for researches and practitioners to choose an appropriate setting (i.e. DNN architecture and attribute encoding) for a given event log.  
To overcome this problem, we conducted an empirical evaluation of three proven DNN architectures and five promising encoding techniques from the ML literature with five real-life event logs.
Academia and practice can benefit from our paper's three contributions. 
Concerning the next activity prediction, our contributions are: 
an overview of existing DL-based PBPM techniques, four findings supporting the design of novel DL-based PBPM techniques using context and directions for future research.
%
%
%
%
%
%
%
%

\section*{Acknowledgments}
This project is funded by the German Federal Ministry of Education and Research (BMBF) within the framework programme \textit{Software Campus}\\ (www.softwarecampus.de) under the number 01IS17045.


\bibliography{main}

\newpage

%
\section*{Appendix}

\begin{table}[H]
\centering

\includegraphics[height=0.93\textheight]{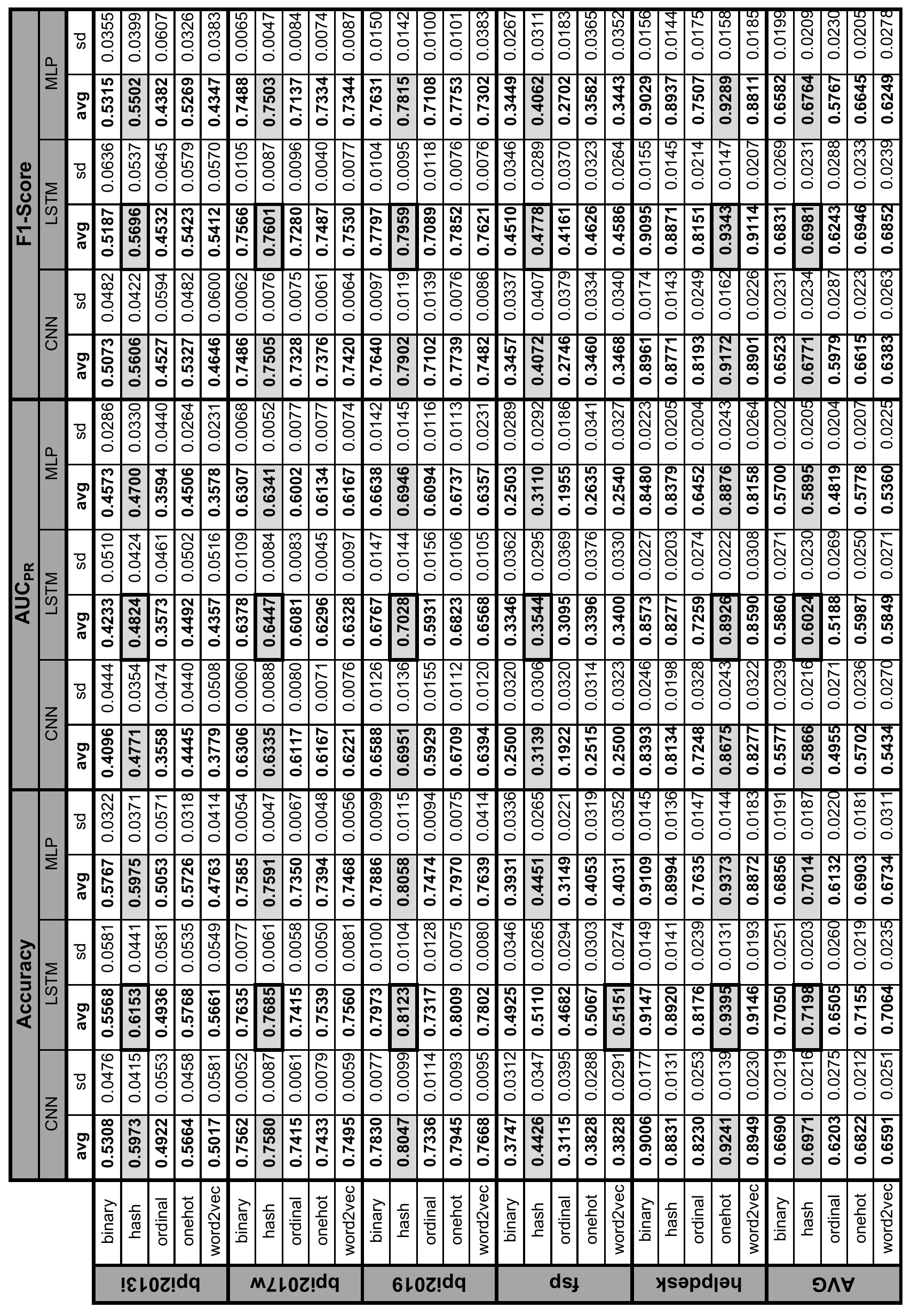}

\caption{Mean and standard deviation over the ten folds.} 

\label{fig:accuracy}

\end{table}


\end{document}